\pdfoutput=1
\documentclass[11pt]{article}
\usepackage[preprint]{acl} % "review", "final": camera-ready, "preprint": non-anonymous with page numbers

\usepackage{times}
\usepackage{latexsym}
\usepackage[T1]{fontenc}
\usepackage[utf8]{inputenc}
\usepackage{microtype}
\usepackage{inconsolata}
\usepackage{graphicx}

\usepackage{amsmath,bm}
\usepackage{enumitem}
\usepackage{booktabs,multirow,multicol,makecell}
\usepackage{subcaption}

\newcommand{\fref}[1]{\hyperref[#1]{Figure~\getrefnumber{#1}}}
\newcommand{\tref}[1]{\hyperref[#1]{Table~\getrefnumber{#1}}}
\newcommand{\sref}[1]{\hyperref[#1]{\S\getrefnumber{#1}}}
\newcommand{\aref}[1]{\ref{#1}}

\newcommand{\ignore}[1]{}

\DeclareMathOperator{\abs}{abs}
\DeclareMathOperator{\acc}{acc}
\DeclareMathOperator{\ncvg}{ncvg}

\title{Domain Pre-training Impact on Representations}

\author{
  {\bf Cesar Gonzalez-Gutierrez}\hspace{1em}
  {\bf Ariadna Quattoni} \\
  Universitat Politècnica de Catalunya, Barcelona, Spain \\
  \texttt{cesar.gonzalez.gutierrez@upc.edu},\\
  \texttt{aquattoni@cs.upc.edu}
}

\begin{document}
\maketitle

\begin{abstract}
This empirical study analyzes the effects of the pre-training corpus on the quality of learned transformer representations. We focus on the representation quality induced solely through pre-training. Our experiments show that pre-training on a small, specialized corpus can yield effective representations, and that the success of combining a generic and a specialized corpus depends on the distributional similarity between the target task and the specialized corpus.
\end{abstract}

\section{Introduction}

Since 2018, pre-training (PT) has become a standard step in model development, demonstrating effective transfer learning for diverse natural language understanding tasks \citep{peters-etal-2018-deep, devlin-etal-2019-bert, radford2018improving}. This approach leverages textual corpora to induce useful representations by minimizing a self-supervised language modeling loss function. This representation is usually leveraged via transfer learning for downstream tasks using supervised data.

\begin{figure}[t]
\begin{subfigure}{\linewidth}
    \centering
    \caption*{Low-Annotation Probing (ALC\%)}\label{fig:heatmap/probing}
    \includegraphics[width=\linewidth]{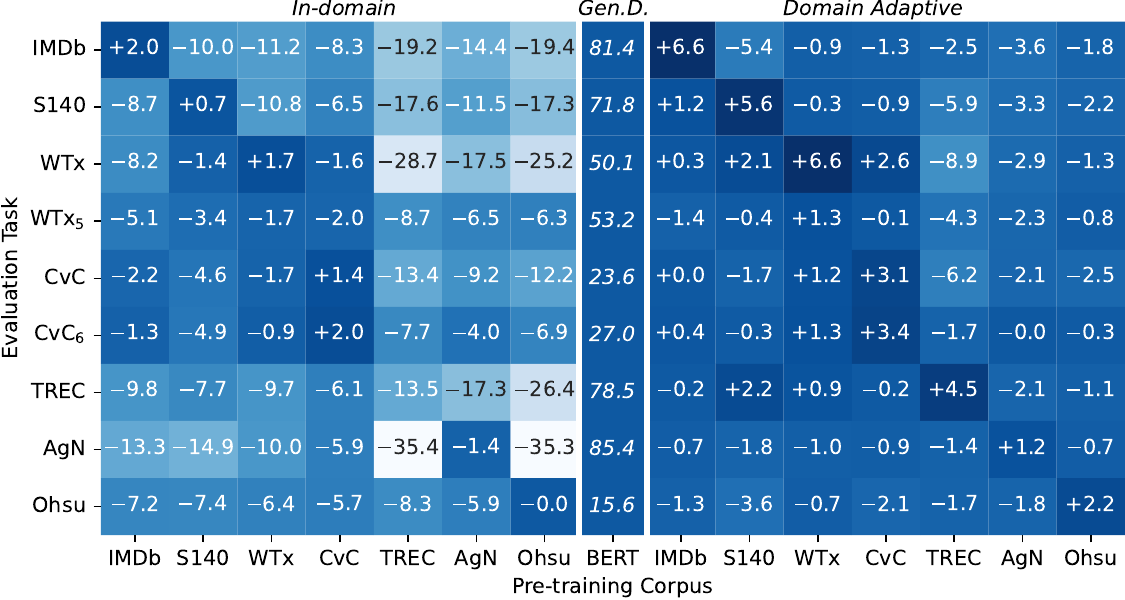}
\end{subfigure}
\begin{subfigure}{\linewidth}
    \centering
    \caption*{Task Alignment (THAS\%)}\label{fig:heatmap/task-alignment}
    \includegraphics[width=\linewidth]{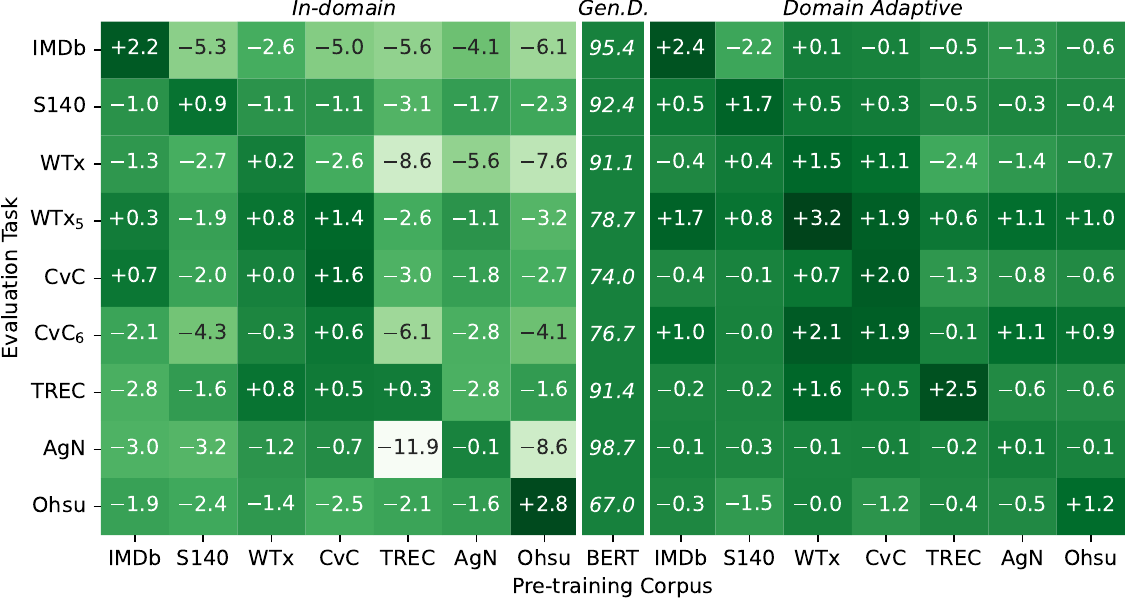}
\end{subfigure}
\caption{Cross-domain representation performance. We report absolute performance for GD embeddings (center), and relative improvement for ID and DA.}
\label{fig:heatmap}
\end{figure}

When selecting a PT corpus, we generally have three options: 1) use a large, generic corpus $\bm{G}$, similar to training a foundational model; 2) use a smaller, \textit{specialized} corpus $\bm{S}$, which is expected to be more relevant to the target task; 3) combine both $\bm{G}$+$\bm{S}$, as in domain-adaptive PT \citep{gururangan-etal-2020-dont}, where a model initially pre-trained on $\bm{G}$ is further refined using $\bm{S}$.

For example, when developing a toxicity filter for a new forum with only a few labeled comments, one option is to use a pre-trained representation computed on a generic corpus $\bm{G}$ (such as BERT). Alternatively, we could construct a specialized corpus $\bm{S}$ by collecting posts from the forum itself or from other related sources. We can then pre-train using either $\bm{G}$ alone or a combination o both $\bm{G}$+$\bm{S}$. We refer to the textual distribution from which a PT corpus is sampled as its \textit{domain}.

The goal of this paper is to understand how pre-training corpus's choice affects the quality of the learned representation. We focus on measuring representation quality after self-supervised pre-training, without subsequent supervised training (as in fine-tuning). Following prior work, we use three representation quality metrics: a standard probing technique \citep{ettinger-etal-2016-probing, adi2017finegrained}, and two label-representation alignment metrics, one based on hierarchical clustering structures \citep{gonzalez-gutierrez-etal-2023-analyzing}, the other one based on data-dependent complexity \citep{yauney-mimno-2021-comparing}.

We conduct an empirical analysis using a wide range of data sources and tasks, evaluating representation quality under different pre-training scenarios. We derive the following conclusions: 1) If the specialized corpus $\bm{S}$ is close to the target task distribution, and it is not too small, pre-training on $\bm{S}$ can be as effective as pre-training on $\bm{G}$. 2) Pre-training on both $\bm{G}$ and $\bm{S}$ (domain-adaptive pre-training) can sometimes improve performance compared to using $\bm{G}$ alone, but it can also lead to decreased performance. 3) The success of the domain-adaptive strategy depends on the similarity between the target task distribution and that of $\bm{S}$. We validate this claim by showing a consistent correlation between distributional similarity and domain-adaptive performance improvement, using two different similarity measures. 
%While this conclusion might seem intuitive, to the best of our knowledge, we are the first ones to substantiate this claim with a thorough empirical study.

\section{Pre-training Effect on Transformer Embeddings}
\label{experiments}

We begin this section with a description of the experimental setting followed by a discussion of the main results.

\subsection{Experimental Setting}
We consider three \textbf{pre-training scenarios}:
\begin{itemize}[topsep=0em,itemsep=0em,leftmargin=1em,after=\vspace{.5em}]
\item In-domain (\textbf{ID}): Pre-training a model from scratch on a domain-specific corpus $\bm{S}$.
\item General domain (\textbf{GD}): The common PLM training approach of using a large generic corpus $\bm{G}$.
\item Domain-adaptive (\textbf{DA}): Starting with a PLM trained on $\bm{G}$, continuing pre-training on $\bm{S}$.
\end{itemize}

To simulate domain-specific pre-training \textbf{corpora} ($\bm{S}_\text{domain}$), we used the unlabeled text in the following classification datasets: IMDb \citep[\ignore{18M words;}][]{maas-etal-2011-learning}, Sentiment 140 \citep[\ignore{21M;}][]{go2009twitter}, Wiki Toxic \citep[\ignore{11M;}][]{wulczyn2017ex}, Civil Comments \citep[\ignore{100M;}][]{borkan2019nuanced}, TREC \citep{li-roth-2002-learning, hovy-etal-2001-toward}, AG-News \citep{zhang2015character}, and Ohsumed \citep{hersh1994ohsumed}. Test partitions are excluded as held-out sets and are used to compute the representation quality metrics. As a general corpus $\bm{G}$, we employ a model pre-trained on BERT corpus (BookCorpus plus English Wikipedia). \tref{tab:corpus} shows PT corpus sizes.

\begin{table}[bt]
\centering
\ignore{
\begin{tabular}{lr}
\toprule
Corpus & \#tokens \\
\midrule
IMDb & 24M \\ % train+dev 7,846,783, train+dev+unsup 23,588,834
Sentiment 140 & 37M \\ % 36,680,959
Wiki Toxic & 15M \\ % clean 12,431,039, orig 15,141,989
Civil Comments & 110M \\ % clean 110,311,922, orig 130,396,952
\cmidrule(lr){1-2}
BookCorpus & 800M \\
English WikiPedia & 2,500M \\
\bottomrule
\end{tabular}
}
\begin{small}
\begin{tabular}{llr}
\toprule
Corpus & Task / Text & \#words \\
\midrule
IMDb & sentiment / movie reviews & 18M \\ % pool 5,844,680 +unsup 17,566,362
Sentiment 140 & sentiment / tweets & 21M \\ % 21,081,841
Wiki Toxic & toxicity / user discussion & 11M \\ % clean 10,979,621 orig 10,734,904
Civil Comments & toxicity / user comments & 100M \\ % clean 99,895,746 orig 97,568,262
TREC & topic / short questions & 56k \\ % 55,636
AG-News & topic / news & 4.5M \\ % 4,541,694
Ohsumed & topic / medical abstracts & 1.8M \\ % 1,801,802
\cmidrule(lr){1-3}
BookCorpus & \hspace{.8em}--\hspace{.8em} / books & 800M \\
Wikipedia (EN) & \hspace{.8em}--\hspace{.8em} / encyclopedia & 2.5B \\
%\multicolumn{2}{l}{\multirow[c]{2}{*}{\makecell[cl]{BERT $\Bigl\{$} \makecell[cl] {BookCorpus \\ English WikiPedia}}} & 800M \\
%& & 2.5B \\
\bottomrule
\end{tabular}
\end{small}
\caption{Pre-training corpora.}
\label{tab:corpus}
\end{table}

We choose a BERT \textbf{model architecture} \citep{devlin-etal-2019-bert} because it is well studied transformer model \citep{rogers-etal-2020-primer} and has a size suitable for running multiple pre-training experiments. Following \citet{liu2019roberta}, we omit the next sentence prediction (NSP) task and focus on masked-LM (MLM). To obtain \textbf{sentence embeddings}, we extract mean token embeddings from the last layer. In \aref{appendix/representations}, we employ other representation functions to explore the roles of different layers and tokens.

To quantify the changes underwent by representations during PT, we employ three \textbf{evaluation metrics} targeted at the model's embedding space quality w.r.t. a task: probing and two label-representation alignment scores. In \aref{appendix/dbi}, we explore an intrinsic clustering quality metric that does not depend on task labels.

Probing \citep{ettinger-etal-2016-probing, adi2017finegrained} uses weak classifiers to evaluate the task performance attributable to the representation. We use low-annotation probes to test the representation's ability to uncover structures that enable learning from few samples.
In particular, the probes are MaxEnt classifiers on top of the embeddings trained with sample sizes ranging from 100 to 1000, increasing in steps of 100. We report the area under this learning curve (ALC).

Task Hierarchical Alignment Score \citep[THAS;][]{gonzalez-gutierrez-etal-2023-analyzing} quantifies the alignment between representation hierarchical clustering structures and task labels. 
This metric measures the degree of cluster \textit{purity} at different hierarchical levels. A representation capable of perfectly separating $n$ classes into $n$ pure clusters will obtain the maximum score. More precisely, we used agglomerative clustering on the embeddings and, for each partition, measured the area under the precision-recall curve using in-cluster class prevalence as label predictions for each data point.

Data-Dependent Complexity \citep[DDC;][]{yauney-mimno-2021-comparing} quantifies the compatibility between a representation and a binary classification task. It captures patterns through the eigen decomposition of a kernel matrix that measures sample similarity in the representation space. Label alignment is evaluated based on the extent to which the label vectors can be reconstructed from their projections onto the top eigenvectors of the kernel matrix.

These three metrics are computed on the test partitions of the \textbf{tasks} associated with the benchmarks described in \sref{tab:corpus}. Additionally, we constructed two new multi-class tasks from the two toxicity benchmarks consisting on predicting the sub-type of toxic comments: Wiki Toxic\textsubscript{5} and Civil Comments\textsubscript{6}.
We repeat each experiment five times using different random seeds, and report average performance. Further experimental details can be found in \aref{appendix/experimental_details}.

\subsection{Main Results}
\label{cross-domain}

\begin{table}[tbp]
\centering
\begin{tabular}{@{\hspace{0em}}l cc cc@{\hspace{0em}}}
\toprule
 & \multicolumn{2}{c}{Binary Tasks} & \multicolumn{2}{c}{All Tasks} \\    
Metric & ncvg & E[acc$_\text{L1}$] & ncvg & E[acc$_\text{L1}$] \\
\midrule
$\Delta$Probe  & 75.85    &  80.30  &   73.91  &   57.18 \\
$\Delta$THAS   & 66.67    &  64.89  &   55.39  &   56.26 \\
$\Delta$DDC    & 71.71    &  96.00  &          &         \\
\bottomrule
\end{tabular}
\caption{Spearman correlation (\%) between similarity of pre-training and target task distributions and representation quality improvement gains.}
\label{tab:correlation}
\end{table}

We study how the choice of pre-training corpus affects representation quality. More specifically, our goal is to understand the necessary conditions for cross-domain generalization. To this end, we first pre-train models on each of the seven corpora $\mathbf{S}_\text{domain}$ described in \tref{tab:corpus}. This is done under two settings: in-domain (i.e., training solely on $\bm{S}_\text{domain}$), and domain-adaptive (i.e., training with both $\bm{G}$+$\bm{S}_\text{domain}$). We then evaluate each representation across all nine target task using the three representation quality metrics: low-annotation probing, task alignment, and DDC. Additionally, we compute the representation quality for the general-domain representation (BERT) as a baseline.

For each quality metric, we generate two matrices, $M_\text{ID}$ and $M_\text{DA}$, corresponding to the in-domain and domain-adaptive settings, respectively. These are shown in \fref{fig:heatmap}, with the DDC matrices reported in \aref{appendix/results}. Each matrix entry $M(i,j)$ represents the performance difference between the evaluated representation and the general-domain baseline. Positive values indicate improvement for low-annotation probing and task alignment, while negative values indicate improvement for DDC (lower is better). For example, $M_\text{DA}(\text{WTx}, \text{CvC})$ denotes the difference in performance evaluated on the $\text{WTx}$ task when using domain-adaptive pre-training on $\bm{S}_\text{CvC}$ corpus, compared to using the general pre-trained representation.

Focusing on the in-domain setting (matrices on the left) and specifically on entries where the pre-training corpus matches the evaluation task domain, we find that in most cases, pre-training on a smaller, specialized dataset yields representations whose quality is comparable to those derived from a general-domain model trained on a much larger corpus. One exception is the TREC task, which can be explained by the limited size of $S_\text{TREC}$.
Still focusing on the domain-matched cases, we observe that the largest improvements in representation quality occur when using the domain-adaptive strategy \citep{gururangan-etal-2020-dont}.

\begin{figure}[t]
\centering
\includegraphics[width=\linewidth]{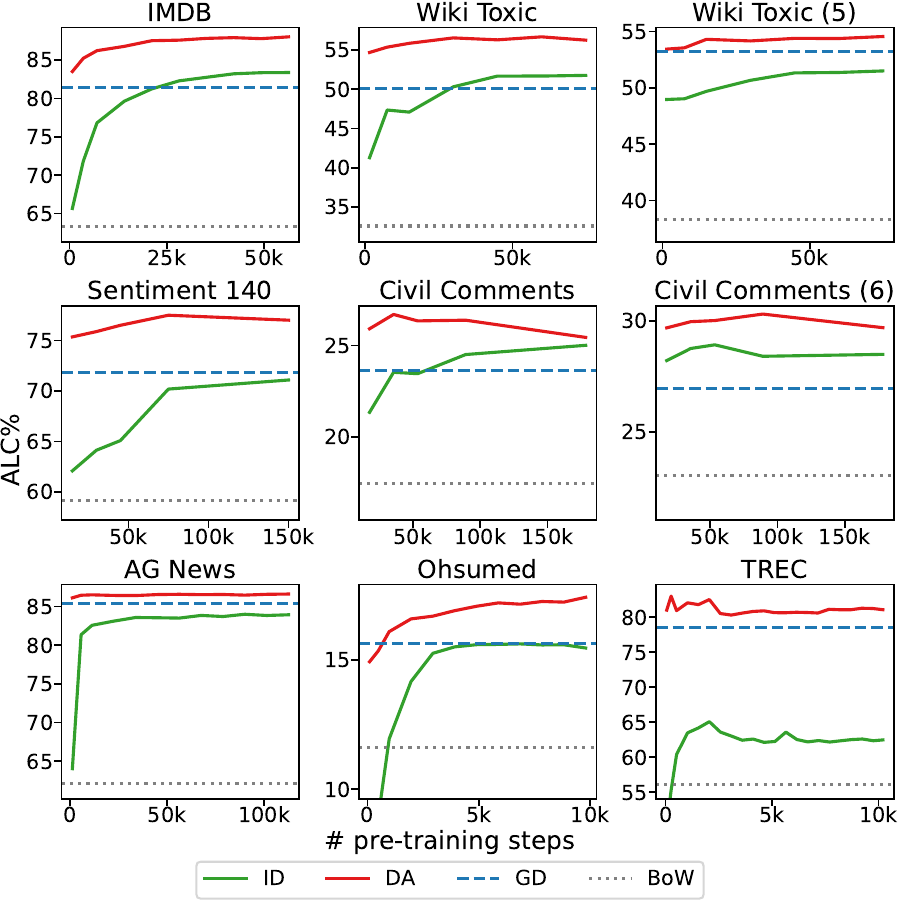}
\caption{Probe performance as a function of PT steps.
%A sparse bag-of-words (BoW) baseline is shown for comparison.
}
\label{fig:lines_probing}
\end{figure}

\fref{fig:lines_probing} complements the domain-matched results showing low-annotation probing performance as a function of the number of PT steps. (\fref{fig:lines_align} shows analogous results for task alignment.) We observe that in-domain PT can achieve performance comparable to general-domain PT, given a sufficient number of pre-training iterations. 

Now, turning our attention to the cross-domain performance, where the pre-training data comes from a domain $\bm{S}_c$ different from the task domain $\bm{S}_t$, the outcomes are mixed. In some cases, domain-specific pre-training improves representation quality; in others, it leads to performance degradation. A natural explanation for this phenomenon is that the effectiveness of domain-specific pre-training depends on the similarity between the distribution of the pre-training data domain and the target task domain.
%\citep{grangier-iter-2022-trade}.

To test this hypothesis, we computed two distributional similarity metrics. The first, $n$-gram coverage ($\ncvg$) measures the proportion of $n$-grams observed in the target task distribution that also appear in the pre-training distribution:
$$ \ncvg = \frac{|N_{t} \cap N_{c}|}{|N_{t}|}, $$
where $N_{t}$ is the set of unique $n$-grams (up to size $n$) observed in the target distribution $\bm{S}_t$, and $N_{c}$ is the analogous set in the pre-training corpus $\bm{S}_c$.

A classical metric for measuring distribution distance is the Kullback-Leibler divergence \citep{kullback1951information} between the $n$-gram distributions $P_{t}$ and $P_{c}$:
$$ KL = \sum_{n \in N_{t}}P_{t}(n) \log \frac{P_{t}(n)}{P_{c}(n)}. $$
However, this metric is problematic in our case due to sparsity: for many $n$-grams, $P_{c}(n)$ will be zero. To address this, we propose a more robust metric, suitable for sparse distributions. This metric estimates the expected L1-accuracy when using $P_{c}(n)$ as a proxy for $P_{t}(n)$:
$$ E[\acc_\text{L1}] = 1 - \sum_{n \in N_{t}}P_{t}(n) \cdot  \abs({P_{t}(n)}-{P_{c}(n)}). $$
Notice that both metrics are asymmetric.

To validate our hypothesis, we computed the correlation between representation quality improvements under the domain-adaptive scenario and these two similarity metrics. As shown in \tref{tab:correlation}, there is a strong correlation between distributional similarity and gains in representation quality. Additional correlation graphs are provided in \aref{appendix/correlation}.

\section{Related Work}

In theoretical studies, \citet{ge2024on, deng2024generalization} attributed the advantage of PT to the induction of useful representations and learning complexity reduction. \citet{tripuraneni2020theory} showed that shared representations enable transfer learning, improving generalization across tasks, even when annotation coverage is sparse \citep{du2021fewshot}. Domain adaptation depends on the diversity of those tasks and the relative sample sizes \citep{grangier-iter-2022-trade}. In this work, we empirically investigate the adaptability of transformer representations and their ability to generalize across diverse tasks in a controlled experimental setting.

Domain-specific data has been used to train different PLMs from scratch \citep[][\textit{inter alia}]{beltagy-etal-2019-scibert, lee2019biobert} or via domain adaptation \cite{gururangan-etal-2020-dont, han-eisenstein-2019-unsupervised}. \citet{aharoni-goldberg-2020-unsupervised} studied the implicit notion of domain in PLMs, \citet{krishna-etal-2023-downstream} studied task-domain PT, and \citet{chronopoulou-etal-2022-efficient} leveraged the cross-domain overlap using adapters. In contrast, we investigate the representation changes that enable domain adaptation in different PT scenarios.

Representation properties have been studied using probing tasks
\citep{ettinger-etal-2016-probing, adi2017finegrained, conneau-etal-2018-cram, hewitt-manning-2019-structural} or analyzing their relation to annotations \citep{gonzalez-gutierrez-etal-2023-analyzing, yauney-mimno-2021-comparing, zhou-srikumar-2021-directprobe}.
Representation learning dynamics has been explored across various syntactic \citep{chiang-etal-2020-pretrained, saphra-lopez-2019-understanding}, semantic \citep{templeton2024scaling, liu-etal-2021-probing-across, liu-etal-2019-linguistic}, or multilingual model capabilities \citep{wang-etal-2024-probing-emergence, blevins-etal-2022-analyzing}. The role of representations in generalization has been studied for linguistic phenomena \citep{choshen-etal-2022-grammar, warstadt-etal-2020-learning} or factual knowledge \citep{zhang-etal-2021-need}. These works have not explored dynamics in varying PT scenarios, with a focus on cross-domain generalization.

\section{Conclusion}

This paper compared pre-trained representations obtained under different PT scenarios. We used three representation quality metrics to evaluate how effectively transformer representations learned from diverse pre-training corpus can be leveraged for a target task. We draw two main conclusions: 1) The success of cross-domain representation transfer can be predicted by the degree of similarity between the $n$-gram distributions of the pre-training and target domains. While this conclusion may seem intuitive, to the best of our knowledge, this is the first empirical study to provide a through analysis of cross-domain adaptation of transformer representations to substantiate it. 2) Pre-trained representations learned from relatively small, domain-specific corpora can be highly competitive. This suggests that the relevance of the PT data may be more important than its size. High-quality models can thus be developed using only domain-specific data, without requiring extensive GD corpora and with a fraction of the computational resources.

\section*{Limitations}

This empirical study explores the properties of transformer-based representations using BERT pre-trained models. While our focus is on any transformer-based representation, we have not compared our results with other transformer architectures. We believe the findings are representative, but a broader experimental setup would allow for more robust conclusions. 

Although current state-of-the-art LLMs present interesting properties that deserve our attention, the methods presented in this work do not scale. In other words, pre-training very large LMs is not feasible with medium-size computational resources.

Fine-tuning is a widely used approach for adapting pre-trained representations for downstream tasks. However, this work focuses solely on the changes induced by pre-training, without any supervised learning. Our aim is to understand how self-supervised pre-training alone shapes representations to support cross-domain transferability. The impact of fine-tuning on representation quality is an important research direction in its own right, which we leave for future work.

The evaluation of embedding performance using the DDC metric is limited to binary classification tasks. While this measure is consistent with the other representation quality metrics used, it may limit generalizability.

\section*{Potential Risks}

We do not foresee any potential societal risks derived from the use of the methods presented in this work.

\anonymize{
\section*{Acknowledgements}
This project has received funding from the European Research Council (ERC) under the European Union's Horizon 2020 research and innovation programme under grant agreement No 853459. The authors gratefully acknowledge the computer resources at ARTEMISA, funded by the European Union ERDF and Comunitat Valenciana as well as the technical support provided by the Instituto de Física Corpuscular, IFIC (CSIC-UV). This research is supported by a recognition 2021SGR-Cat (01266 LQMC) from AGAUR (Generalitat de Catalunya).
}

\bibliography{anthology,custom}

\appendix

\section{Appendix}

\subsection{Experimental Details}
\label{appendix/experimental_details}

\paragraph{Models}
Our experiments employ the BERT implementation from the HuggingFace Transformers library \citep{wolf-etal-2020-transformers} with PyTorch \citep{ansel2024pytorch} backend. The models were trained using single Tesla Volta V100 32GiB PCIe GPUs.

MaxEnt probes are implemented using Scikit-learn toolbox \citep{pedregosa2011scikitlearn} and NumPy \citep{harris2020array}.

\paragraph{Text Corpora}
Domain-specific corpora are constructed concatenating the text fields within a dataset, if they contained multiple. We used all the partitions present in the dataset except the test set. This includes training, development (if available), and unsupervised (IMDb only) partitions.

Toxicity datasets (Wiki Toxic and Civil Comments) are pre-processed to remove any markup and non-alphanumeric characters except relevant punctuation.

\paragraph{Evaluation Datasets}

The tasks used to evaluate representation quality are summarized in \tref{tab:datasets}.

\begin{table}[htbp]
\centering
\begin{small}
\begin{tabular}{@{\hspace{0em}} l @{\hspace{1em}}l @{\hspace{0.5em}} r @{\hspace{1em}} r @{\hspace{1em}} r  @{\hspace{1em}} r @{\hspace{0em}}}
\toprule
Dataset & Task & $|\mathcal{Y}|$ & Prior & len. & train / test \\
\midrule
IMDb & sentiment & 2 & 0.5 & 233 & 25k / 25k \\
Sentiment 140 & sentiment & 2 & 0.5 & 14 & 1.6M / 498 \\
Wiki Toxic & toxicity & 2 & 0.096 & 68 & 160k / 64k \\
Wiki Toxic (5) & toxicity & 5 & imb. & 52 & 9.8k / 3.4k \\
Civ. Com. & toxicity & 2 & 0.08 & 53 & 1.9M / 97k \\
Civ. Com. (6) & toxicity & 6 & imb. & 49 & 145k / 7.4k \\
TREC & topic & 6 & imb. & 10 & 5.5k / 500 \\
AG-News & topic & 4 & $1/|\mathcal{Y}|$ & 38 & 120k / 7.6k \\
Ohsumed & topic & 23 & imb. & 175 & 10k / 13k \\
\bottomrule
\end{tabular}
\end{small}
\caption{Evaluation dataset statistics.}
\label{tab:datasets}
\end{table}

\paragraph{Evaluation Metrics}
For low-annotation probing (ALC), we report accuracy for the binary balanced datasets and multi-class datasets, as is standard practice. For the imbalanced binary datasets (Civil Comments and Wiki Toxic), we report F1 of the target class. 

In \aref{appendix/curves}, the curves from which ALC and THAS aggregate metrics where computed can be found in \fref{fig:curves}.

Following \citet{yauney-mimno-2021-comparing}, DDC score is computed as the ratio of the real annotation's DDC to the average DDC over random annotations.

\paragraph{Pre-training}

\tref{tab:params_pt} presents the parameters used in model pre-training.

\begin{table}[htbp]
\centering
\begin{tabular}{lr}
\toprule
Parameter & Value \\
\midrule
architecture & BERT\textsubscript{BASE} \\
hidden size & 768 \\
max. tokens & 512 \\
vocabulary size & 30,522 \\
activation & gelu \\
dropout & 0.1 \\
batch size & 96 \\
optimizer & AdamW \\
learning rate & 5e-5 \\
weight decay & linear \\
mixed precision & fp16 \\
\bottomrule
\end{tabular}
\caption{Model pre-training parameters.}
\label{tab:params_pt}
\end{table}

We pre-trained BERT models for different total number of updates depending on the size of the dataset (see \fref{fig:lines_align}).
\citet{devlin-etal-2019-bert} pre-trained BERT for 1M update steps, approximately 40 epochs over the 3.3B word corpus. In comparison, the training length and computation resources needed for our models is orders of magnitude smaller.
\anonymize{}
\footnote{
For reproducibility, we will release the weights of the domain pre-trained models after publication.
}

\subsection{Supplementary Results}
\label{appendix/results}

\begin{figure}[htbp]
\centering
\small{Data-Dependent Complexity}
\includegraphics[width=.9\linewidth]{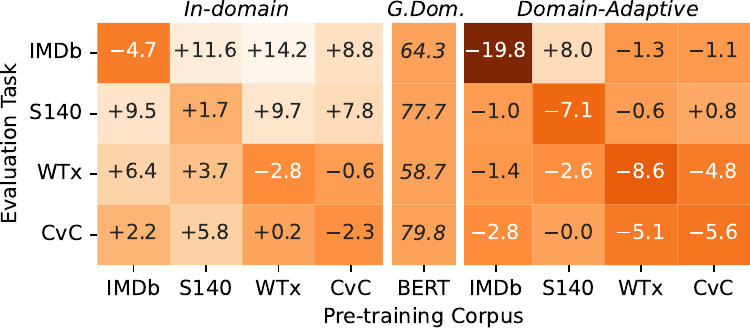}
\caption{Cross-domain representation DDC performance using binary tasks.}
\label{fig:heatmap/ddc}
\end{figure}

\fref{fig:heatmap/ddc} shows the data-dependent complexity matrices corresponding to the analysis performed in \sref{cross-domain}.

\begin{figure}[htbp]
\centering
\includegraphics[width=\linewidth]{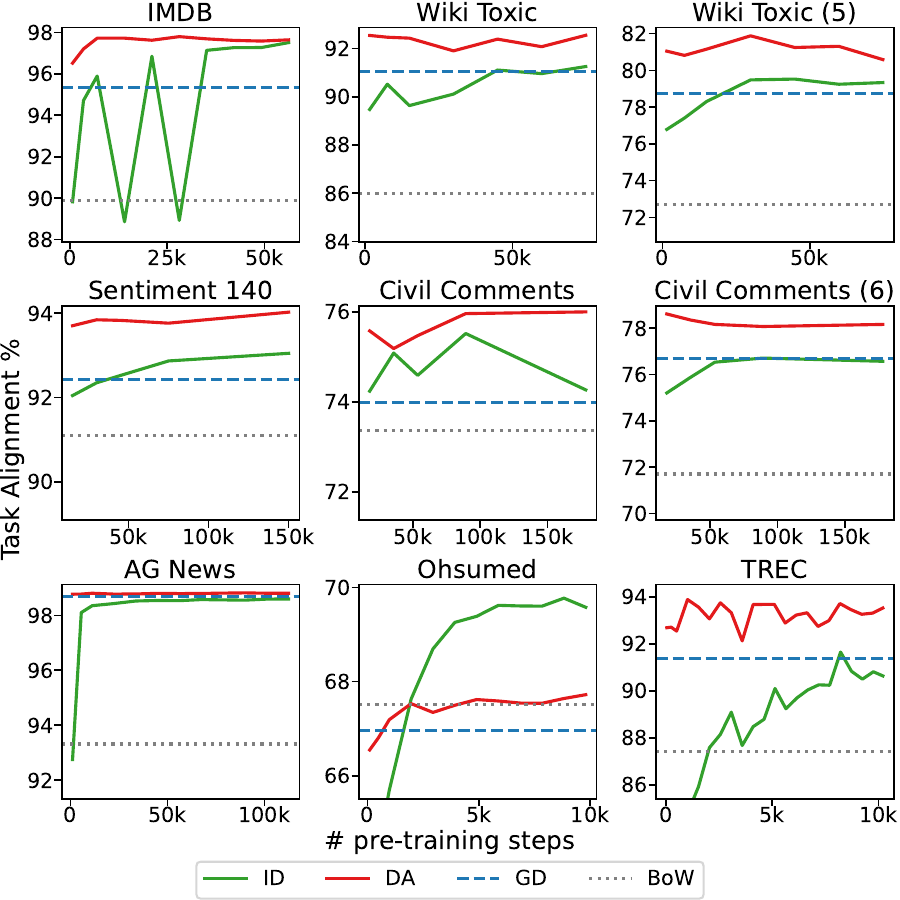}
\caption{Task alignment (THAS\%) performance as a function of PT updates in the three PT scenarios. A BoW baseline is shown for comparison.}
\label{fig:lines_align}
\end{figure}

\fref{fig:lines_align} shows the pre-training dynamic curves using the task alignment metric, as in \sref{cross-domain}.

\subsection{Performance Improvements Correlation}
\label{appendix/correlation}

\begin{figure}[htbp]
\centering
\includegraphics[width=\linewidth]{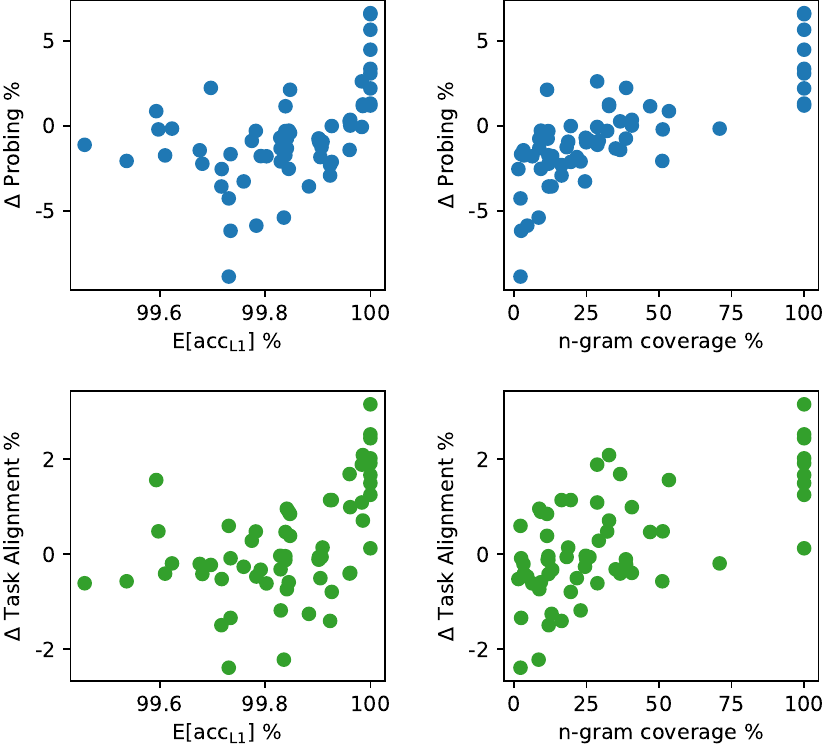}
\caption{Correlation between similarity of pre-training corpus and target task distributions and representation quality differences using all datasets.}
\label{fig:correlation_all}
\end{figure}

\begin{figure}[htbp]
\centering
\includegraphics[width=\linewidth]{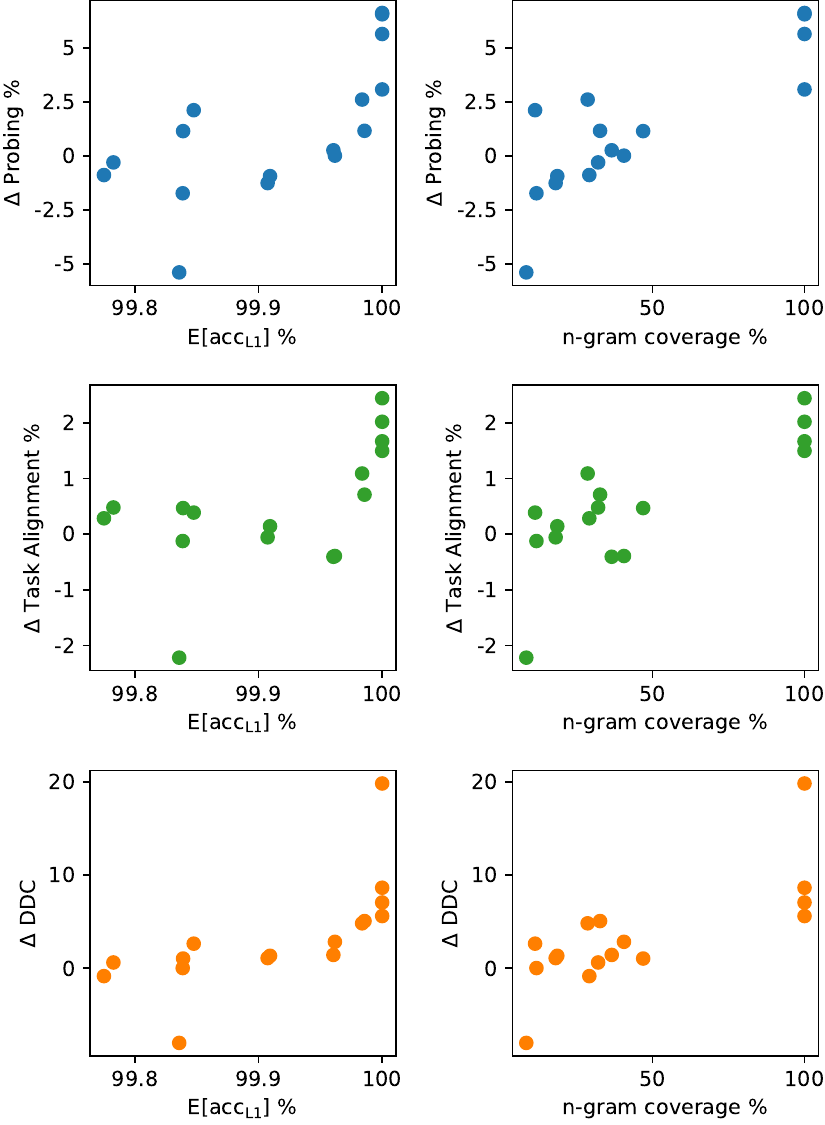}
\caption{Correlation between distribution similarity and representation quality improvements for binary datasets.}
\label{fig:correlation_bin}
\end{figure}

To complement the correlation study presented in \sref{cross-domain}, this section presents scatter plots between the distribution distance measurements and representation quality improvements. This comparison only considers the representations obtained in the domain-adaptive setting. \fref{fig:correlation_all} presents the correlation plot for all the tasks, whereas \fref{fig:correlation_bin} presents analogous plots for the binary datasets (DDC is only defined for binary datasets).

\subsection{Clustering Quality}
\label{appendix/dbi}

To analyze the changes in the embedding structures produced by pre-training from the clustering quality perspective, we computed DBI scores at various granularity levels. We evaluated the hierarchical clustering partitions employed to compute task alignment. Due to the computational cost of DBI, we only computed scores every 25 partitions.

In \fref{fig:curves} (right), we report DBI curves as a function of the number of clusters in the partitions for the same representations considered in the task alignment curves. Lower DBI scores indicate better clustering partitions. \tref{tab:pt-dbi} shows the ADBI aggregate scores, i.e. the area under the DBI curves, corresponding to the eight feature extraction strategies presented in \aref{appendix/representations}.

Domain-adaptive PT has a limited impact on clustering quality. Interestingly, for coarse partitions with fewer clusters (early points in the curves), clustering quality often declines relative to the GD baseline.
In contrast, in-domain PT alters the cluster structures leading to larger differences in DBI scores compared to the baseline. This is expected, as the clustering structures, starting from a random spatial distribution, evolve as PT progresses.

In general, PT decreases the quality of clusters. For a given representation, whether using in-domain PT or domain-adaptive PT, DBI scores tend to worsen as the number of training epochs increases.
This illustrates how the spatial encoding that embeddings undergo during PT does not translate into compactness and separability of the induced structures.
Instead, the induced structures do not lend themselves to straightforward assumptions about the global structure of the embedding space.

The behavior of this embedding space property is uncorrelated with task alignment and probing, as the relative ranking of representations is not preserved. This observation is consistent with the findings in \citet{gonzalez-gutierrez-etal-2023-analyzing}.

\subsection{Curves for Aggregate Evaluation Metrics}
\label{appendix/curves}

\begin{figure*}[tbp]
{\hfill \hspace{1em} \textit{Task Alignment} \hspace{5em} \textit{Low-annotation Probing} \hspace{4em} \textit{Clustering Quality}\hfill}\vspace{1em}
\centering
\begin{subfigure}[t]{\textwidth}
    \centering
    \includegraphics[width=.33\textwidth]{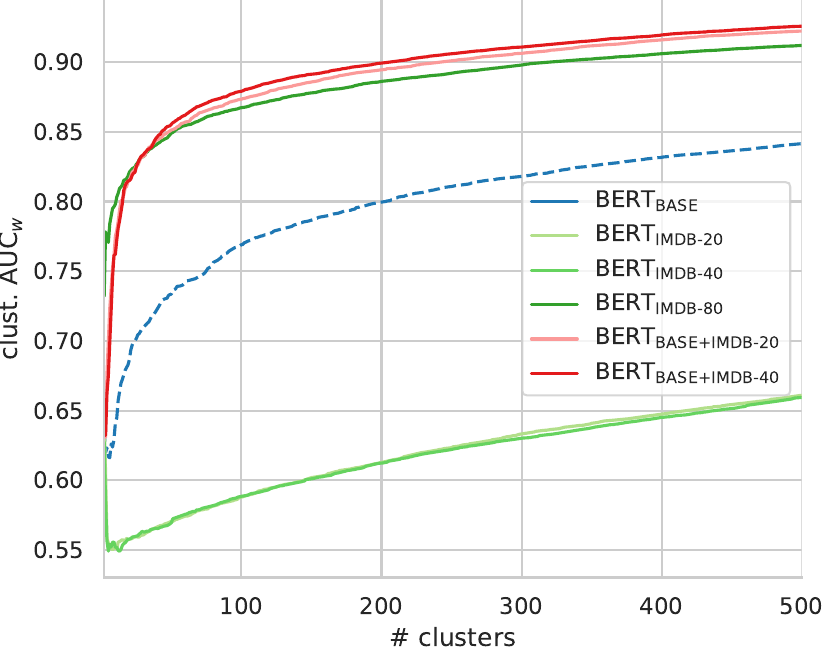}
    \includegraphics[width=.32\textwidth]{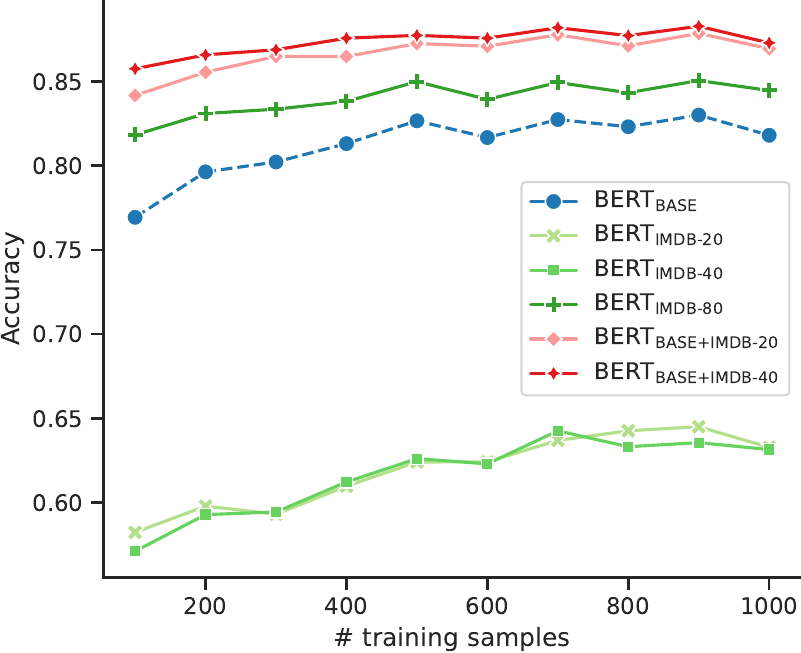}
    \includegraphics[width=.33\linewidth]{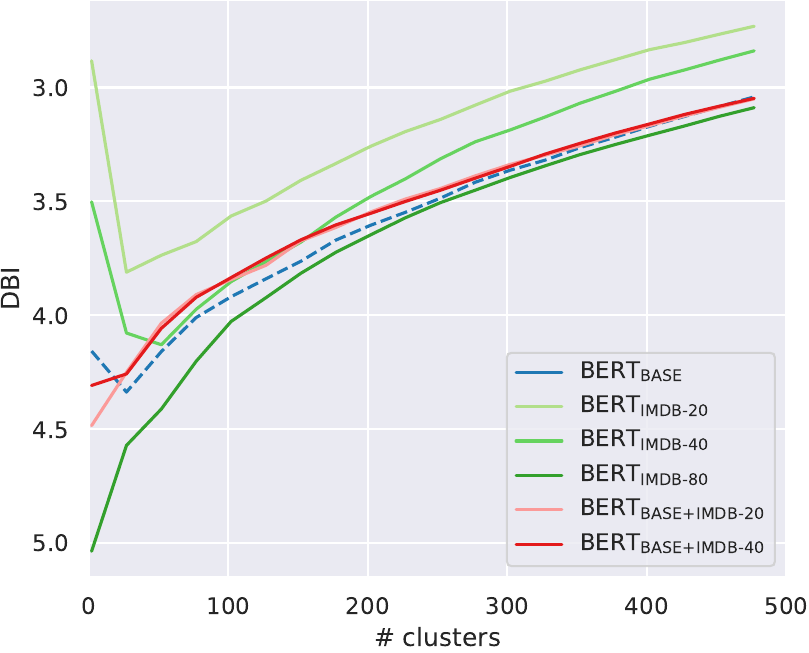}
    \caption{IMDb}
    \label{fig:pt-imdb}
\end{subfigure}\vspace{1em}
\begin{subfigure}[t]{\textwidth}
    \centering
    \includegraphics[width=.33\textwidth]{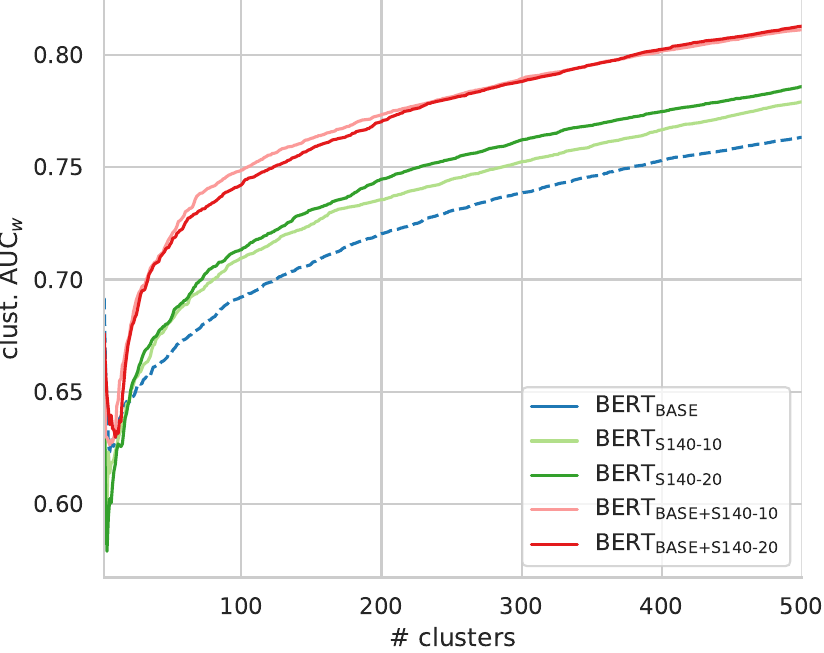}
    \includegraphics[width=.32\textwidth]{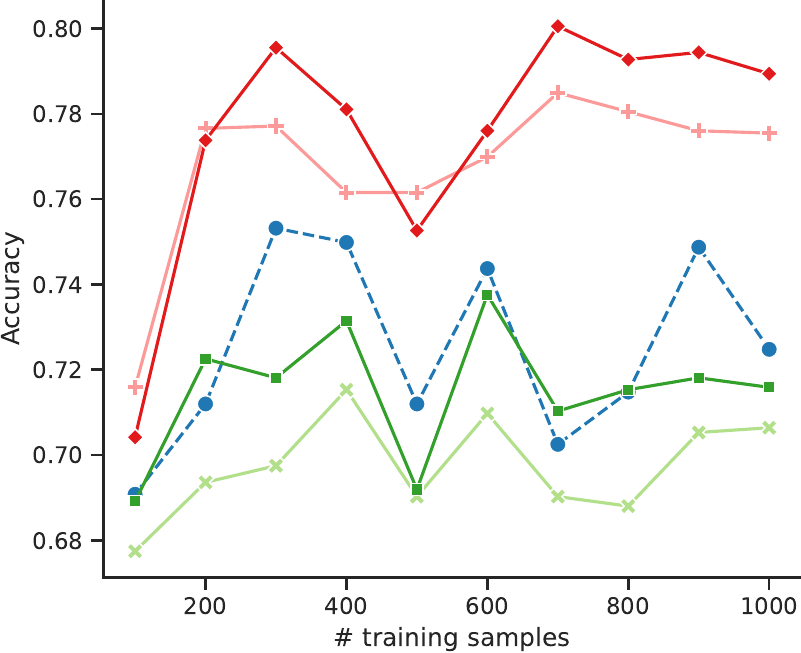}
    \includegraphics[width=.33\linewidth]{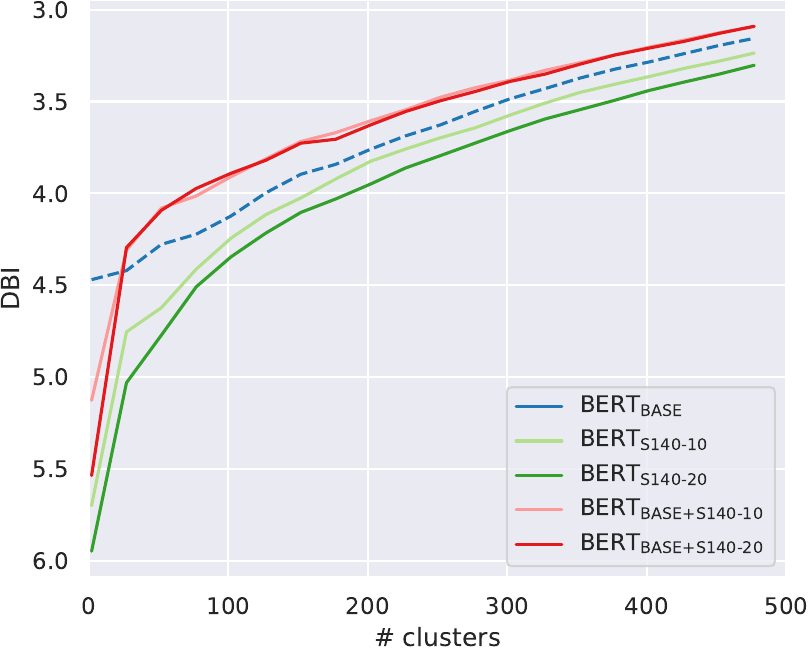}
    \caption{Sentiment140}
    \label{fig:pt-s140}
\end{subfigure}\vspace{1em}
\begin{subfigure}[t]{\textwidth}
    \centering
    \includegraphics[width=.33\textwidth]{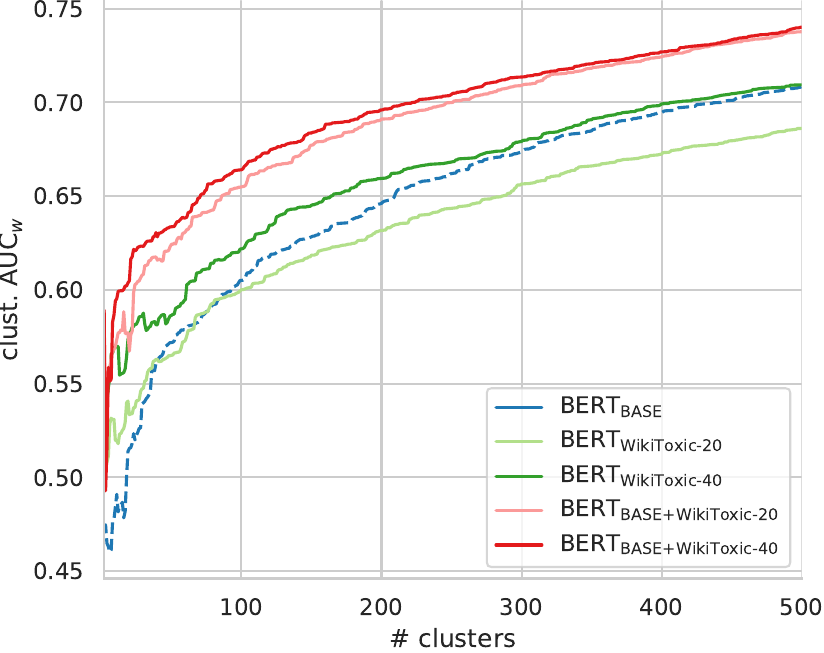}
    \includegraphics[width=.32\textwidth]{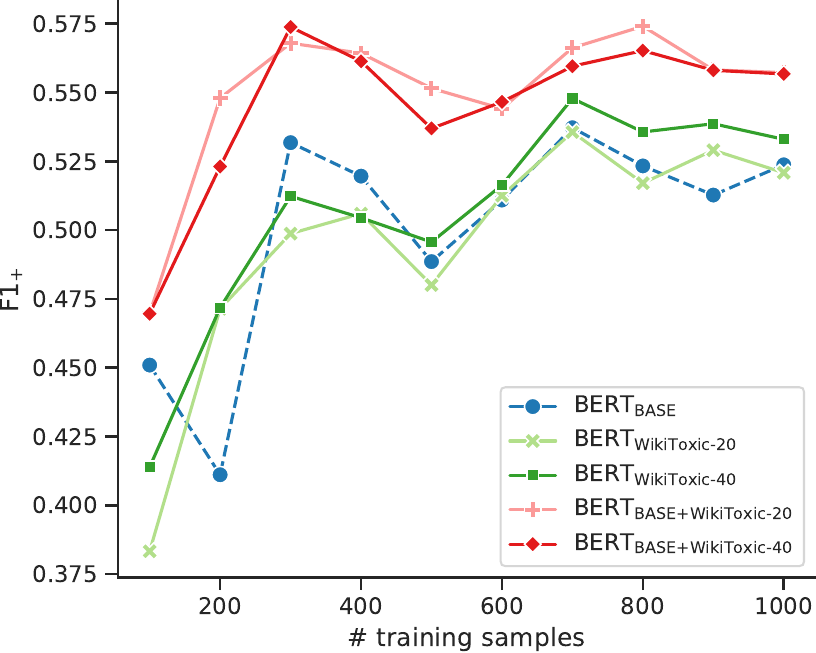}
    \includegraphics[width=.33\linewidth]{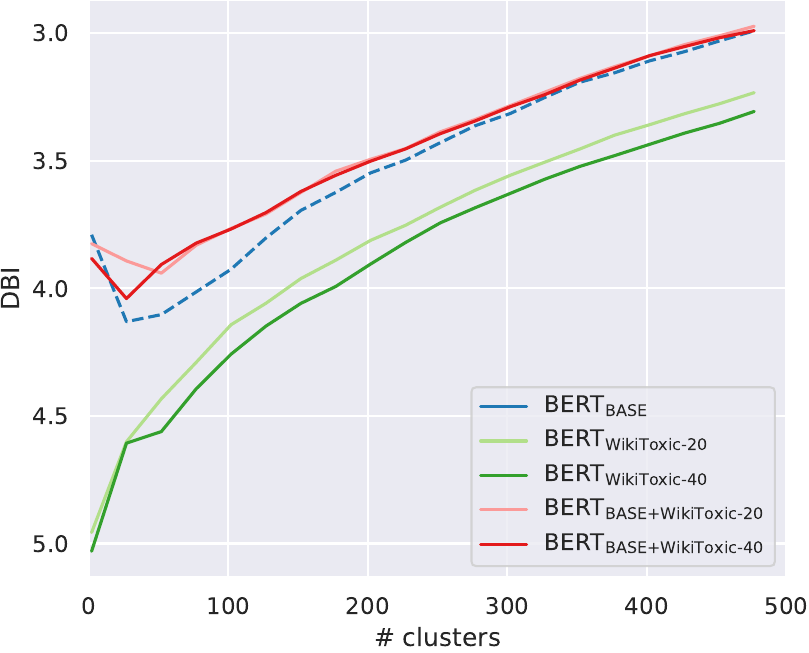}
    \caption{Wiki Toxic}
    \label{fig:pt-wikitoxic}
\end{subfigure}\vspace{1em}
\begin{subfigure}[t]{\textwidth}
    \centering
    \includegraphics[width=.33\textwidth]{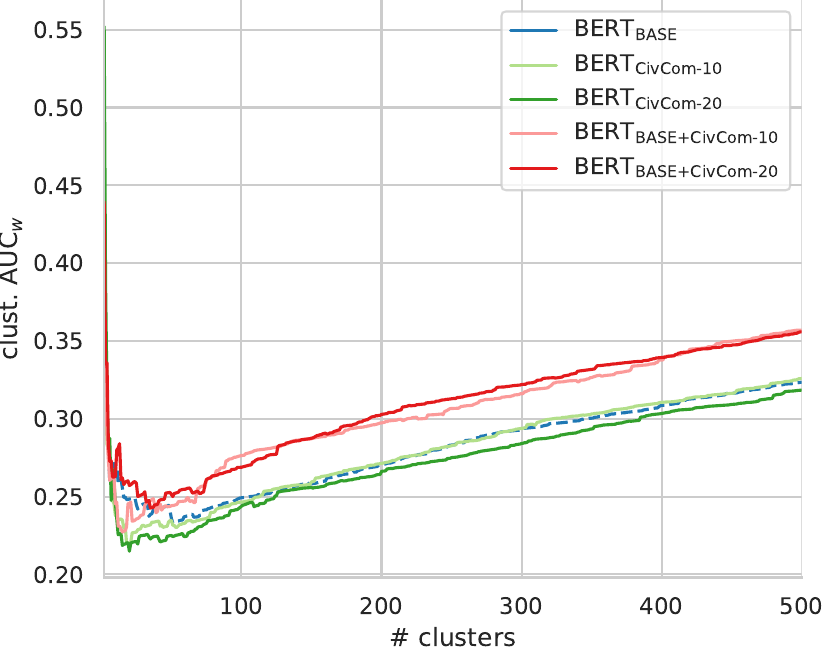}
    \includegraphics[width=.32\textwidth]{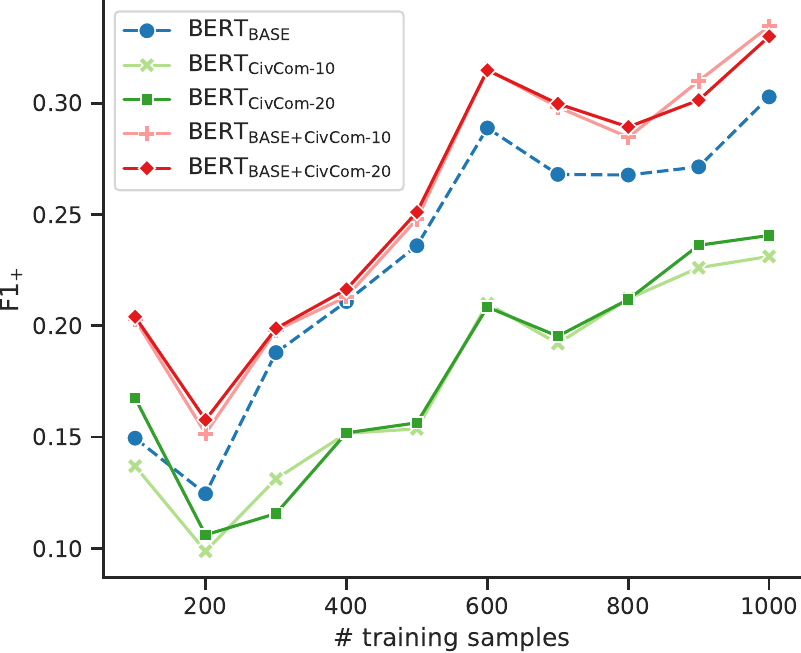}
    \includegraphics[width=.33\linewidth]{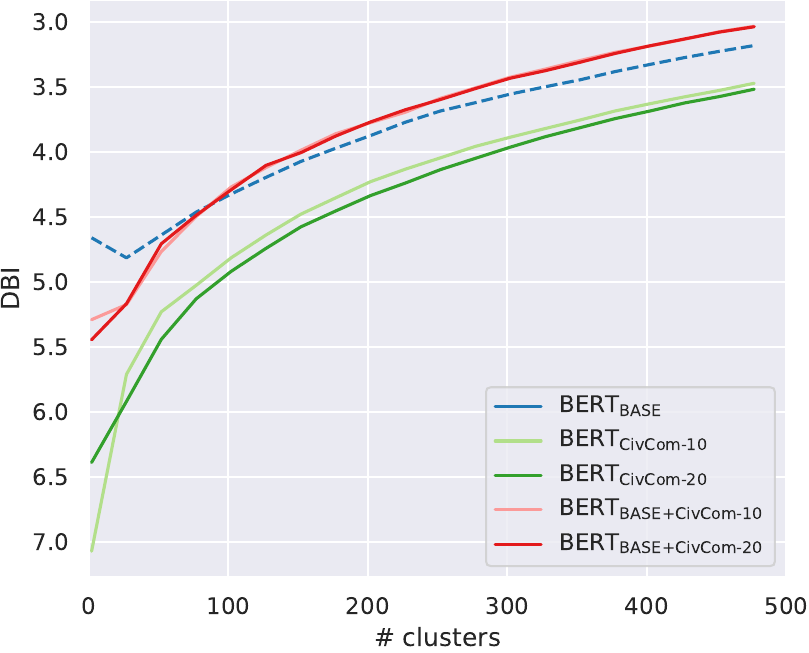}
    \caption{Civil Comments}
    \label{fig:pt-civcom}
\end{subfigure}
\caption{
Representation performance curves for different datasets using BERT embeddings generated from the last layer with token average pooling. Representations are produced under three PT scenarios: general domain (dashed blue), in-domain (green), and domain-adaptive (red). Multiple pre-training stages are shown with the number of epochs indicated in the representation name.
}
\label{fig:curves}
\end{figure*}

\fref{fig:curves} illustrates the curves used to calculate the aggregate metrics presented in \sref{experiments} and \sref{appendix/dbi}. The left of the figure shows the first 500 points of the task alignment curve used to compute THAS. The center of the figure presents the low-annotation probing learning curves used to compute ALC. To the right, DBI curves as a function of cluster granularity define the clustering quality aggregate metric.

For these examples, we used pre-trained text representations using a single feature extraction strategy (the last layer with token average pooling) of our four binary classification tasks.
We denote BERT pre-trained with the standard corpus as BERT\textsubscript{BASE}. Models pre-trained in-domain are denoted by the dataset as a subscript (e.g., BERT\textsubscript{IMDb} for a model pre-trained solely on IMDb). Domain-adapted models indicate both corpora in subscripts, as in BERT\textsubscript{BASE+S140}, which represents BERT\textsubscript{BASE} continued with Sentiment140. Epoch counts are included in the model notation (e.g., BERT\textsubscript{IMDb-80} for the 80-epoch IMDb model).

\subsection{Comparison of Feature Extraction Strategies}
\label{appendix/representations}

To better understand the role of model's layers and tokens during PT, Tables \ref{tab:pt-align}, \ref{tab:alc} and \ref{tab:pt-dbi} present aggregate metrics for the same datasets and models as in \sref{experiments}, evaluated using different feature extraction strategies.

We consider four layer extraction methods: last layer ($_1$), second-to-last ($_2$), concatenating the last four ($_\text{cat1:4}$) and averaging all twelve ($_{\mu1:12}$); in combination to two token pooling strategies: average of all tokens ($^\mu$), and taking the \texttt{[CLS]} token alone ($^\text{CLS}$).

Interestingly, the conclusions in \sref{experiments} remain consistent across the various representation functions, indicating that embedding improvement during PT affects all the layer and token representations in the model.

Analyzing BERT's token representations reveals that the general domain BERT achieves its strongest performance when extracting embeddings from all the tokens, outperforming the representations derived from the \texttt{[CLS]} token, regardless of the layer.
This pattern remains consistent for ID representations, with the exception of models pre-trained on Wiki Toxic.
This outcome is expected, as the NSP objective, which primarily leverages the \texttt{[CLS]} token, was not computed during PT.
Interestingly, DA models generate relatively strong \texttt{[CLS]} token representations. This result aligns with the findings of \citet{liu2019roberta}, who suggests that MLM is sufficient for effective PT.

Regarding layer selection, feature extraction typically favors upper layers \citep{devlin-etal-2019-bert, reimers-gurevych-2019-sentence}, as they are more specialized than lower layers \citep{ethayarajh-2019-contextual}. From the perspective of our metrics, this holds true in these PT scenarios, although the differences between layers are not very pronounced. Strategies that incorporate lower layers, such as averaging all 12, often produce strong representations and even outperform upper-layer strategies for datasets like Civil Comments and Wiki Toxic.

\begin{table*}[htbp]
\centering
\begin{small}
\begin{tabular}{lll rrrrrrrr}
\toprule
 &  &  & \multicolumn{8}{c}{Task Alignment \%} \\
\cmidrule{4-11}
Dataset & PT & Ep. & $_1^{\mu}$ & $_1^{\text{CLS}}$ & $_2^{\mu}$ & $_2^{\text{CLS}}$ & $_\text{cat1:4}^{\mu}$ & $_\text{cat1:4}^{\text{CLS}}$ & $_{\mu1:12}^{\mu}$ & $_{\mu1:12}^{\text{CLS}}$ \\
\midrule
\multirow[t]{6}{*}{IMDb} & \multirow[t]{3}{*}{ID} & 20 & 88.86 & 87.72 & 88.82 & 87.34 & 88.86 & 87.45 & 88.71 & 87.49 \\
 &  & 40 & 88.93 & 87.38 & 88.72 & 86.99 & 88.83 & 87.06 & 88.78 & 87.21 \\
 &  & 80 & 97.51 & 95.74 & 97.50 & 95.20 & 97.10 & 94.94 & 97.04 & 94.84 \\
\cmidrule(l){2-11}
 & GD &  & 95.36 & 94.56 & 95.31 & 93.45 & 95.25 & 93.43 & 94.58 & 92.49 \\
\cmidrule(l){2-11}
 & \multirow[t]{2}{*}{DA} & 20 & 97.72 & 98.70 & 97.78 & 98.95 & 97.71 & 98.89 & 96.37 & 98.51 \\
 &  & 40 & 97.80 & 99.04 & 97.90 & 99.20 & 97.79 & 99.15 & 96.55 & 98.87 \\
\cmidrule{1-11}
\multirow[t]{5}{*}{Sentiment140} & \multirow[t]{2}{*}{ID} & 10 & 93.05 & 92.79 & 93.01 & 92.54 & 92.87 & 92.73 & 92.63 & 92.60 \\
 &  & 20 & 93.20 & 92.66 & 93.23 & 92.72 & 93.12 & 92.71 & 92.83 & 92.52 \\
\cmidrule(l){2-11}
 & GD &  & 92.43 & 92.36 & 92.40 & 92.10 & 92.68 & 91.86 & 92.20 & 91.34 \\
\cmidrule(l){2-11}
 & \multirow[t]{2}{*}{DA} & 10 & 94.02 & 94.87 & 94.04 & 94.92 & 94.18 & 94.80 & 93.40 & 94.18 \\
 &  & 20 & 94.10 & 94.84 & 94.13 & 95.00 & 94.32 & 94.93 & 93.45 & 94.25 \\
\cmidrule{1-11}
\multirow[t]{5}{*}{Wiki Toxic} & \multirow[t]{2}{*}{ID} & 20 & 90.11 & 90.60 & 90.12 & 90.69 & 90.50 & 90.59 & 90.76 & 89.59 \\
 &  & 40 & 90.96 & 91.86 & 90.99 & 92.11 & 91.00 & 92.01 & 91.07 & 91.19 \\
\cmidrule(l){2-11}
 & GD &  & 91.06 & 86.54 & 90.49 & 86.64 & 90.75 & 86.72 & 91.07 & 87.26 \\
\cmidrule(l){2-11}
 & \multirow[t]{2}{*}{DA} & 20 & 91.91 & 91.09 & 91.90 & 92.33 & 91.96 & 91.80 & 91.85 & 91.93 \\
 &  & 40 & 92.08 & 91.12 & 91.91 & 92.63 & 92.18 & 91.62 & 92.02 & 91.75 \\
\cmidrule{1-11}
\multirow[t]{5}{*}{Civil Comments} & \multirow[t]{2}{*}{ID} & 10 & 74.27 & 74.25 & 74.33 & 72.25 & 73.89 & 74.55 & 74.29 & 73.68 \\
 &  & 20 & 73.96 & 74.20 & 73.77 & 72.68 & 74.22 & 74.23 & 74.31 & 74.05 \\
\cmidrule(l){2-11}
 & GD & & 73.98 & 71.33 & 73.23 & 71.46 & 73.69 & 71.28 & 74.21 & 71.87 \\
\cmidrule(l){2-11}
 & \multirow[t]{2}{*}{DA} & 10 & 76.00 & 76.69 & 75.93 & 75.63 & 75.61 & 76.75 & 75.43 & 74.92 \\
 &  & 20 & 75.93 & 76.86 & 75.85 & 75.26 & 75.49 & 77.21 & 75.51 & 75.32 \\ 
\bottomrule
\end{tabular}
\end{small}
\caption{Task alignment of pre-trained BERT embeddings across different datasets and feature extraction strategies.}
\label{tab:pt-align}
\end{table*}

\begin{table*}[htbp]
\centering
\begin{small}
\begin{tabular}{lll rrrrrrrr}
\toprule
 &  &  & \multicolumn{8}{c}{ALC \%} \\
\cmidrule{4-11}
Dataset & PT & Ep. & $_1^{\mu}$ & $_1^{\texttt{CLS}}$ & $_2^{\mu}$ & $_2^{\texttt{CLS}}$ & $_\text{cat1:4}^{\mu}$ & $_\text{cat1:4}^{\texttt{CLS}}$ & $_{\mu1:12}^{\mu}$ & $_{\mu1:12}^{\texttt{CLS}}$ \\
\midrule
\multirow[t]{6}{*}{\makecell[t]{IMDb \\ (acc)}} & \multirow[t]{3}{*}{ID} & 20 & 61.60 & 55.34 & 61.89 & 54.27 & 62.17 & 55.41 & 63.15 & 56.01 \\
 &  & 40 & 61.70 & 55.11 & 61.63 & 54.12 & 62.62 & 55.10 & 62.99 & 55.91 \\
 &  & 80 & 84.11 & 81.33 & 83.98 & 80.55 & 83.86 & 80.99 & 83.04 & 80.53 \\
\cmidrule{2-11}
 & GD & & 81.39 & 78.52 & 81.23 & 77.09 & 82.14 & 78.69 & 80.15 & 76.61 \\
\cmidrule{2-11}
 & \multirow[t]{2}{*}{DA} & 20 & 86.72 & 88.03 & 86.67 & 88.51 & 87.04 & 89.03 & 84.92 & 87.87 \\
 &  & 40 & 87.12 & 89.40 & 87.35 & 89.30 & 87.57 & 89.73 & 85.42 & 89.10 \\
\cmidrule{1-11}
\multirow[t]{5}{*}{\makecell[t]{Sentiment140 \\ (acc)}} & \multirow[t]{2}{*}{ID} & 10 & 70.19 & 68.73 & 69.74 & 68.59 & 70.90 & 70.25 & 69.63 & 68.49 \\
 &  & 20 & 71.92 & 70.41 & 71.50 & 70.04 & 72.35 & 71.97 & 71.44 & 71.05 \\
\cmidrule{2-11}
 & GD & & 71.84 & 70.97 & 72.52 & 69.38 & 73.58 & 71.86 & 71.00 & 69.18 \\
\cmidrule{2-11}
 & \multirow[t]{2}{*}{DA} & 10 & 76.71 & 77.09 & 76.80 & 75.89 & 78.47 & 78.31 & 76.29 & 77.24 \\
 &  & 20 & 77.48 & 77.16 & 77.60 & 76.53 & 78.86 & 79.06 & 77.30 & 77.52 \\
\cmidrule{1-11}
\multirow[t]{5}{*}{\makecell[t]{Wiki Toxic \\ (F1$_+$)}} & \multirow[t]{2}{*}{ID} & 20 & 49.96 & 52.96 & 49.54 & 53.21 & 49.63 & 53.11 & 49.17 & 52.81 \\
 &  & 40 & 50.54 & 55.75 & 50.70 & 56.42 & 50.97 & 56.13 & 50.37 & 56.31 \\
\cmidrule{2-11}
 & GD & & 50.09 & 48.18 & 50.10 & 46.58 & 51.41 & 48.11 & 52.60 & 48.49 \\
\cmidrule{2-11}
 & \multirow[t]{2}{*}{DA} & 20 & 54.55 & 54.00 & 55.01 & 54.98 & 54.81 & 55.72 & 55.43 & 54.50 \\
 &  & 40 & 54.68 & 53.75 & 54.51 & 54.76 & 55.05 & 55.51 & 55.39 & 54.34 \\
\cmidrule{1-11}
\multirow[t]{5}{*}{\makecell[t]{Civil Comments \\ (F1$_+$)}} & \multirow[t]{2}{*}{ID} & 10 & 18.78 & 20.44 & 17.44 & 20.29 & 19.01 & 20.26 & 19.84 & 20.67 \\
 &  & 20 & 18.38 & 20.23 & 17.90 & 20.40 & 19.22 & 19.76 & 19.79 & 20.40 \\
\cmidrule{2-11}
 & GD & & 23.63 & 19.63 & 23.08 & 17.52 & 23.92 & 17.88 & 24.61 & 19.43 \\
\cmidrule{2-11}
 & \multirow[t]{2}{*}{DA} & 10 & 25.98 & 26.32 & 25.56 & 26.04 & 26.99 & 26.30 & 26.96 & 27.36 \\
 &  & 20 & 25.94 & 26.16 & 25.64 & 26.18 & 26.88 & 26.15 & 26.12 & 27.69 \\
\bottomrule
\end{tabular}
\end{small}
\caption{Area under the learning curve (ALC) of low-annotation probes for different feature extraction strategies. We report accuracy for balanced datasets and F1 of the target class for imbalanced datasets.}
\label{tab:alc}
\end{table*}

\begin{table*}[htbp]
\centering
\begin{small}
\begin{tabular}{lll rrrrrrrr}
\toprule
 &  &  & \multicolumn{8}{c}{ADBI} \\
 \cmidrule{4-11}
Dataset & PT & Ep. & $_1^{\mu}$ & $_1^{\text{CLS}}$ & $_2^{\mu}$ & $_2^{\text{CLS}}$ & $_\text{cat1:4}^{\mu}$ & $_\text{cat1:4}^{\text{CLS}}$ & $_{\mu1:12}^{\mu}$ & $_{\mu1:12}^{\text{CLS}}$ \\
\midrule
\multirow[t]{6}{*}{IMDb} & \multirow[t]{3}{*}{ID} & 20 & 1.18 & 1.07 & 1.18 & 0.97 & 1.16 & 0.99 & 1.12 & 1.01 \\
 &  & 40 & 1.21 & 0.90 & 1.21 & 0.83 & 1.20 & 0.84 & 1.15 & 0.87 \\
 &  & 80 & 1.39 & 0.92 & 1.38 & 0.88 & 1.30 & 0.89 & 1.25 & 0.93 \\
 \cmidrule{2-11}
 & GD & & 1.35 & 1.36 & 1.32 & 1.32 & 1.31 & 1.35 & 1.29 & 1.33 \\
 \cmidrule{2-11}
 & \multirow[t]{2}{*}{DA} & 20 & 1.34 & 1.33 & 1.34 & 1.33 & 1.32 & 1.37 & 1.29 & 1.35 \\
 &  & 40 & 1.34 & 1.33 & 1.34 & 1.32 & 1.32 & 1.36 & 1.28 & 1.35 \\
\cmidrule{1-11}
\multirow[t]{5}{*}{Sentiment 140} & \multirow[t]{2}{*}{ID} & 10 & 1.39 & 1.26 & 1.41 & 1.28 & 1.38 & 1.28 & 1.37 & 1.33 \\
 &  & 20 & 1.41 & 1.27 & 1.43 & 1.29 & 1.40 & 1.29 & 1.39 & 1.33 \\
\cmidrule{2-11}
 & GD & & 1.36 & 1.27 & 1.35 & 1.23 & 1.36 & 1.26 & 1.36 & 1.25 \\
 \cmidrule{2-11}
 & \multirow[t]{2}{*}{DA} & 10 & 1.36 & 1.24 & 1.36 & 1.25 & 1.36 & 1.24 & 1.35 & 1.19 \\
 &  & 20 & 1.37 & 1.24 & 1.36 & 1.26 & 1.37 & 1.25 & 1.35 & 1.20 \\
\cmidrule{1-11}
\multirow[t]{5}{*}{Wiki Toxic} & \multirow[t]{2}{*}{ID} & 20 & 1.36 & 1.27 & 1.38 & 1.27 & 1.18 & 1.28 & 1.14 & 1.23 \\
 &  & 40 & 1.40 & 1.29 & 1.40 & 1.28 & 1.20 & 1.30 & 1.15 & 1.23 \\
 \cmidrule{2-11}
 & GD & & 1.20 & 1.24 & 1.23 & 1.26 & 1.21 & 1.29 & 1.16 & 1.25 \\
 \cmidrule{2-11}
 & \multirow[t]{2}{*}{DA} & 20 & 1.26 & 1.23 & 1.24 & 1.26 & 1.23 & 1.26 & 1.16 & 1.24 \\
 &  & 40 & 1.27 & 1.25 & 1.25 & 1.26 & 1.23 & 1.27 & 1.16 & 1.25 \\
\cmidrule{1-11}
\multirow[t]{5}{*}{Civil Comments} & \multirow[t]{2}{*}{ID} & 10 & 1.46 & 1.36 & 1.49 & 1.32 & 1.42 & 1.36 & 1.44 & 1.40 \\
 &  & 20 & 1.48 & 1.39 & 1.50 & 1.34 & 1.44 & 1.39 & 1.45 & 1.40 \\
 \cmidrule{2-11}
 & GD & & 1.28 & 1.32 & 1.31 & 1.32 & 1.30 & 1.36 & 1.24 & 1.31 \\
 \cmidrule{2-11}
 & \multirow[t]{2}{*}{DA} & 10 & 1.32 & 1.33 & 1.26 & 1.36 & 1.32 & 1.35 & 1.25 & 1.33 \\
 &  & 20 & 1.32 & 1.35 & 1.26 & 1.37 & 1.32 & 1.37 & 1.25 & 1.34 \\
\bottomrule
\end{tabular}
\end{small}
\caption{Clustering quality measured by the area under the DBI scores for pre-trained embeddings using diverse feature extraction strategies. Lower scores mean better clustering.}
\label{tab:pt-dbi}
\end{table*}

\end{document}